\documentclass[11pt,a4paper]{article}
\usepackage[hyperref]{eacl2021}
\usepackage{times}
\usepackage{latexsym}

\usepackage{url}

\usepackage{mathtools}
\usepackage{mathrsfs}
\usepackage{amssymb}
\usepackage[ruled,linesnumbered]{algorithm2e}
\SetKwRepeat{Do}{do}{while}
\usepackage{multirow}

\usepackage{microtype}

\aclfinalcopy % Uncomment this line for the final submission
\def\aclpaperid{8} %  Enter the acl Paper ID here

\title{Unsupervised Sentence-embeddings by Manifold Approximation and Projection}

\author{Subhradeep Kayal \\
  Prosus N.V. \\
  Amsterdam, Netherlands. \\
  {\tt deep.kayal@pm.me} %\\\And
  }

\date{}

\begin{document}
\maketitle
\begin{abstract}
The concept of unsupervised \emph{universal sentence encoders} has gained traction recently, wherein pre-trained models generate effective task-agnostic fixed-dimensional representations for phrases, sentences and paragraphs. Such methods are of varying complexity, from simple weighted-averages of word vectors to complex language-models based on bidirectional transformers. In this work we propose a novel technique to generate sentence-embeddings in an unsupervised fashion by projecting the sentences onto a fixed-dimensional manifold with the objective of preserving local neighbourhoods in the original space. To delineate such neighbourhoods we experiment with several set-distance metrics, including the recently proposed \emph{Word Mover's distance}, while the fixed-dimensional projection is achieved by employing a scalable and efficient manifold approximation method rooted in topological data analysis. We test our approach, which we term \emph{EMAP} or \emph{Embeddings by Manifold Approximation and Projection}, on six publicly available text-classification datasets of varying size and complexity. Empirical results show that our method consistently performs similar to or better than several alternative state-of-the-art approaches.
\end{abstract}

\section{Introduction}
\label{sec:intro}

\subsection{On sentence-embeddings}
Dense vector representation of words, or \emph{word-embeddings}, form the backbone of most modern NLP applications and can be constructed using context-free \cite{bengio13,w2v,pennington} or contextualized methods \cite{peters,devlin}.

Given that practical systems often benefit from having representations for sentences and documents, in addition to word-embeddings \cite{7389336,yan-etal-2016-docchat}, a simple trick is to use the weighted average over some or all of the embeddings of words in a sentence or document. Although \emph{sentence-embeddings} constructed this way often lose information because of the disregard for word-order during averaging, they have been found to be surprisingly performant \cite{aldarmaki-diab-2018-evaluation}.

More sophisticated methods focus on jointly learning the embeddings of sentences and words using models similar to \emph{Word2Vec} \cite{doc2vec,chen2017efficient}, using encoder-decoder approaches that reconstruct the surrounding sentences of an encoded passage \cite{skipthought}, or training bi-directional LSTM models on large external datasets \cite{infersent}. Meaningful sentence-embeddings have also been constructed by fine-tuning pre-trained bidirectional transformers \cite{devlin} using a Siamese architecture \cite{reimers-gurevych-2019-sentence}.

In parallel to the approaches mentioned above, a stream of methods have emerged recently which exploit the inherent geometric properties of the structure of sentences, by treating them as sets or sequences of word-embeddings. For example, Arora et al. \shortcite{sif} propose the construction of sentence-embeddings based on weighted word-embedding averages with the removal of the dominant singular vector, while R{\"u}ckl{\'e} et al. \shortcite{pmean} produce sentence-embeddings by concatenating several power-means of word-embeddings corresponding to a sentence. Very recently, spectral decomposition techniques were used to create sentence-embeddings, which produced state-of-the-art results when used in concatenation with averaging \cite{kayal-tsatsaronis-2019-eigensent,almarwani-etal-2019-efficient}.

Our work is most related to that of Wu et al. \shortcite{wmemb} who use Random Features \cite{randfeat} to learn document embeddings which preserve the properties of an explicitly-defined kernel based on the Word Mover's Distance \cite{WMDist}. Where Wu et al. predefine the nature of the kernel, our proposed approach can learn the similarity-preserving manifold for a given set-distance metric, offering increased flexibility.

\subsection{Motivation and contributions}
A simple way to form sentence-embeddings is to compute the dimension-wise arithmetic mean of the embeddings of the words in a particular sentence. Even though this approach incurs information loss by disregarding the fact that sentences are sequences (or, at the very least, sets) of word vectors, it works well in practice. This already provides an indication that there is more information in the sentences to be exploited.

Kusner et al. \shortcite{WMDist} aim to use more of the information available in a sentence by representing sentences as a weighted point cloud of embedded words. Rooted in transportation theory, their \emph{Word Mover's distance (WMD)} is the minimum amount of distance that the embedded words of a sentence need to travel to reach the embedded words of another sentence. The approach achieves state-of-the-art results for sentence classification when combined with a \begin{math} k \end{math}-NN classifier \cite{knn}. Since their work, other distance metrics have been suggested \cite{DBLP:conf/iclr/SinghHDJ19,DBLP:journals/corr/abs-1904-10294}, also motivated by how transportation problems are solved.

Considering that sentences are sets of word vectors, a large variety of methods exist in literature that can be used to calculate the distance between two sets, in addition to the ones based on transport theory. Thus, as a \emph{first contribution}, we compare alternative metrics to measure distances between sentences. The metrics we suggest, namely the \emph{Hausdorff distance} and the \emph{Energy distance}, are intuitive to explain and reasonably fast to calculate. The choice of these particular distances are motivated by their differing origins and their general usefulness in the respective application domains.

Once calculated, these distances can be used in conjunction with \begin{math} k \end{math}-nearest neighbours for classification tasks, and \begin{math} k \end{math}-means for clustering tasks. However, these learning algorithms are rather simplistic and the state-of-the-art machine learning algorithms require a fixed-length feature representation as input to them. Moreover, having fixed-length representations for sentences (\emph{sentence-embeddings}) also provides a large degree of flexibility for downstream tasks, as compared to having only relative distances between them. With this as motivation, the \emph{second contribution} of this work is to produce sentence-embeddings that approximately preserve the topological properties of the original sentence space. We propose to do so using an efficient scalable manifold-learning algorithm termed \emph{UMAP} \cite{2018arXivUMAP} from topological data analysis. Empirical results show that this process yields sentence-embeddings that deliver near state-of-the-art classification performance with a simple classifier.

\section{Methodology}
\label{sec:methodology}

\subsection{Calculating distances}
\label{distance}

In this work, we experiment with three different distance measures to determine the distance between sentences. The first measure (Energy distance) is motivated by a useful linkage criterion from hierarchical clustering \cite{hac}, while the second one (Hausdorff distance) is an important metric from algebraic topology that has been successfully used in document indexing \cite{10.1145/2396761.2398499}. The final metric (Word Mover's distance) is a recent extension of an existing distance measure between distributions, that is particularly suited for use with word-embeddings \cite{WMDist}.

Prior to defining the distances that have been used in this work, we first proceed to outline the notations that we will be using to describe them.

\subsubsection{Notations}

Let \begin{math} \mathscr{W} \in \mathbb{R}^{N\times d} \end{math} denote a word-embedding matrix, such that the vocabulary corresponding to it consists of $N$ words, and each word in it, \begin{math} w_i \in \mathbb{R}^d \end{math}, is \begin{math} d \end{math}-dimensional. This word-embedding matrix and its constituent words may come from pre-trained representations such as Word2Vec \cite{w2v} or GloVe \cite{pennington}, in which case \begin{math} d = 300 \end{math}.

Let \begin{math} \mathscr{S} \end{math} be a set of sentences and \begin{math} s, s' \end{math} be two sentences from this set. Each such sentence can be viewed as a set of word-embeddings, \begin{math} \{w\} \in s \end{math}. Additionally, let the length of a sentence, \begin{math} s \end{math}, be denoted as \begin{math} |s| \end{math}, and the cardinality of the set, \begin{math} \mathscr{S} \end{math}, be denoted by \begin{math} |\mathscr{S}| \end{math}.

Let \begin{math} e(w_i, w_j) \end{math} denote the distance between two word-embeddings, \begin{math} w_i, w_j \end{math}. In the context of this paper, this distance is Euclidean:
\begin{equation}
    e(w_i, w_j) = {\|w_i - w_j\|}_2
    \label{euclid}
\end{equation}

Finally, \begin{math} D(s, s') \end{math} denotes the distance between two sentences.

\subsubsection{Energy distance}
\emph{Energy distance} is a statistical distance between probability distributions, based on the inter and intra-distribution variance, that satisfies all the criteria of being a metric \cite{energydist}.

Using the notations defined earlier, we write it as:
\begin{equation}
\begin{gathered}
    D(s, s') = \frac{2}{|s||s'|} \sum_{w_i \in s} \sum_{w_j \in s'} e(w_i, w_j)\\ - \frac{1}{{|s|}^2} \sum_{w_i \in s} \sum_{w_j \in s} e(w_i, w_j)\\ - \frac{1}{{|s'|}^2} \sum_{w_i \in s'} \sum_{w_j \in s'} e(w_i, w_j)
\end{gathered}
\label{ed}
\end{equation}

The original conception of the energy distance was inspired by gravitational potential energy of celestial objects. Looking closely at Equation \ref{ed}, it can be quickly observed that it has two parts: the first term resembles the attraction or repulsion between two objects (or in our case, sentences), while the second and the third term indicate the self-coherence of the respective objects. As shown by Sz\'ekely and Rizzo \shortcite{energydist}, energy distance is scale equivariant, which would make it sensitive to contextual changes in sentences, and therefore make it useful in NLP applications.

\subsubsection{Hausdorff distance}

Given two subsets of a metric space, the \emph{Hausdorff distance} is the maximum distance of the points in one subset to the nearest point in the other. A significant work has gone into making it fast to calculate \cite{hausdorff} so that it can be applied to real-world problems, such as shape-matching in computer vision \cite{hausdorffcv}.

To calculate it, the distance between each point from one set and the closest point from the other set is determined first. Then, the Hausdorff distance is calculated as the maximal point-wise distance. Considering sentences \begin{math} \{s, s'\} \end{math} as subsets of word-embedding space, \begin{math} \mathbb{R}^{d\times N} \end{math}, the \emph{directed} Hausdorff distance can be given as:
\begin{equation}
    h(s, s') = \max_{w_i \in s} \min_{w_j \in s'} e(w_i, w_j)
\end{equation}
such that the symmetric Hausdorff distance is:
\begin{equation}
    D(s, s') = \max \{ h(s,s'), h(s',s) \}
\label{hdist}
\end{equation}

\subsubsection{Word Mover's distance}

In addition to the representation of a sentence as a set of word-embeddings, a sentence \begin{math} s \end{math} can also be represented as a \begin{math} N \end{math}-dimensional normalized term-frequency vector, where \begin{math} n_i^{s} \end{math} is the number of times word \begin{math} w_i \end{math} occurs in sentence \begin{math} s \end{math} normalized by the total number of words in \begin{math} s \end{math}:
\begin{equation}
    n_i^s = \frac{c_i^s}{\sum_{k=1}^{k=N} c_k^s}
\end{equation}
where, \begin{math} c_i^s \end{math} is the number of times word \begin{math} w_i \end{math} appears in sentence \begin{math} s \end{math}.

The goal of the \emph{Word Mover's distance (WMD)} \cite{WMDist} is to construct a sentence similarity metric based on the distances between the individual words within each sentence, given by Equation \ref{euclid}. In order to calculate the distance between two sentences, WMD introduces a transport matrix, \begin{math} T \in \mathbb{R}^{N \times N} \end{math}, such that each element in it, \begin{math} T_{ij} \end{math}, denotes how much of \begin{math} n_i^s \end{math} should be transported to \begin{math} n_j^{s'} \end{math}. Then, the WMD between two sentences is given as the solution of the following minimization problem:
\begin{equation}
\centering
\begin{gathered}
    D(s,s') = \min_{T \geq 0} \sum_{i,j = 1}^{N} T_{ij} e(i,j) \\ \text{subject to,} \enskip \sum_{j = 1}^N T_{ij} = n_i^s \enskip \text{and} \enskip \sum_{i = 1}^N T_{ij} = n_j^{s'}
\end{gathered}
\label{wmdeq}
\end{equation}
Thus, WMD between two sentences is defined as the minimum distance required to transport the words from one sentence to another.

\subsection{Generating neighbourhood-preserving embeddings via non-linear manifold-learning}
\label{umap-sec}

In this work, we propose to construct sentence-embeddings which preserve the neighbourhood around sentences delineated by the relative distances between them. We posit that preserving the local neighbourhoods will serve as a proxy for preserving the original topological properties.

In order to learn a topology-preserving fixed-dimensional manifold, we seek inspiration from methods in non-linear dimensionality-reduction \cite{nldr} and topological data analysis literature \cite{tda}. When broadly categorized, these techniques consist of methods, such as \emph{Locally Linear Embedding} \cite{lle}, that preserve local distances between points, or those like \emph{Stochastic Neighbour Embedding} \cite{sne,tsne} that preserve the conditional probabilities of points being neighbours. However, existing manifold-learning algorithms suffer from two shortcomings: they are computationally expensive and are often restricted in the number of output dimensions. In our work we use a method termed \emph{Uniform Manifold Approximation and Projection (UMAP)} \cite{2018arXivUMAP}, which is scalable and has no computational restrictions on the output embedding dimension.

The building block of UMAP is a particular type of a simplicial complex, known as the Vietoris-Rips complex. Recalling that a \emph{k-simplex} is a \begin{math} k \end{math}-dimensional polytope which is the convex hull of its \emph{k} + 1 vertices, and a simplicial complex is a set of simplices of various orders, the Vietoris-Rips simplicial complex is a collection of 0 and 1-simplices. In essence, this is a means to building a simple neighbourhood graph by connecting the original data points.

\begin{figure}[!htbp]
\centering
    \includegraphics[width=0.48\textwidth]{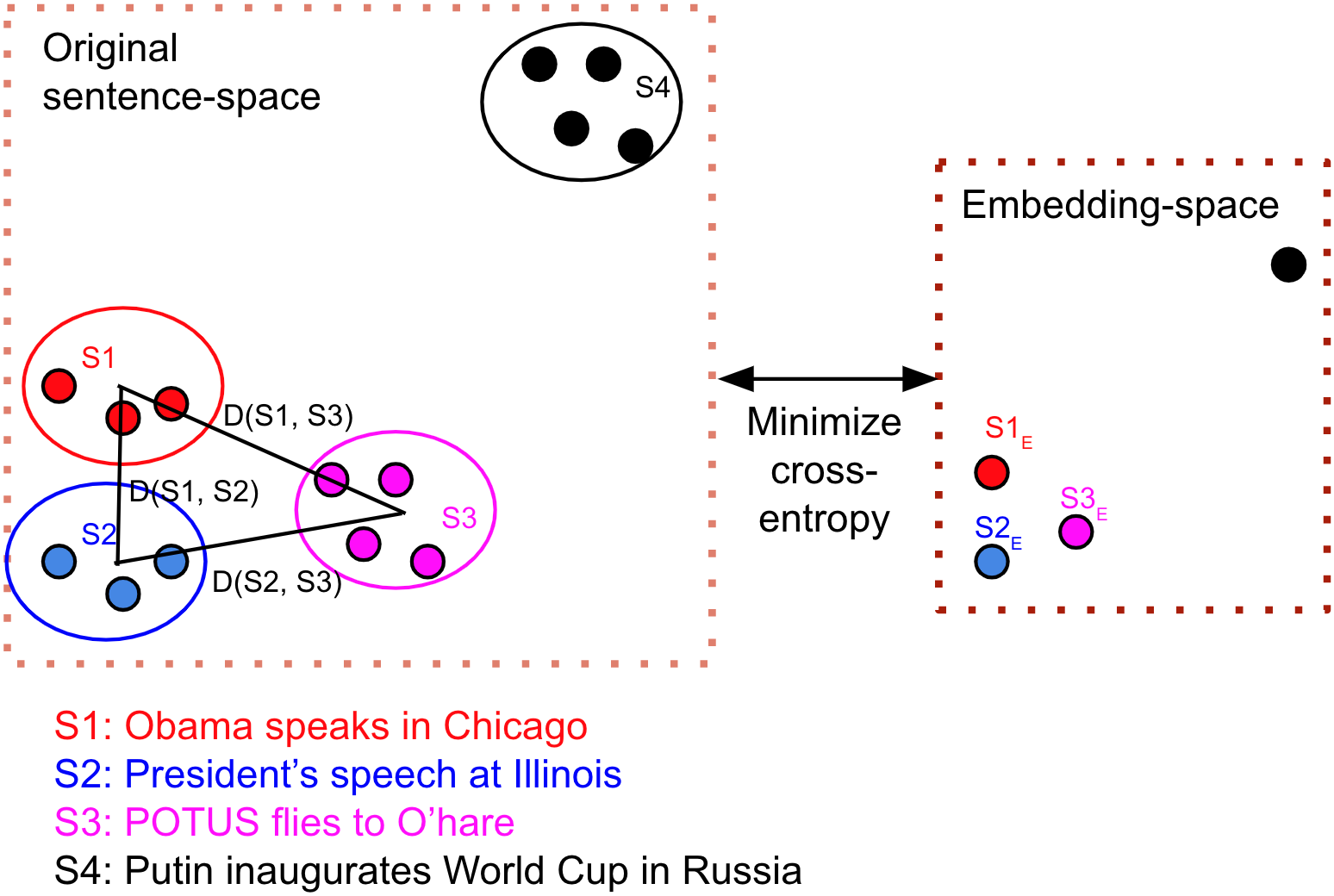}
    \caption{Figure showing a simple example of the embedding algorithm. On the left is the original sentence-space, approximated by the nearest neighbours graph formed by the Vietoris-Rips complex. Instead of points and edges, our simplicial complex has sets of points and edges between them, formed by one of the distance metrics mentioned in Section \ref{distance}. In this example, four sentences, denoted by \begin{math} S1 \end{math} through \begin{math} S4 \end{math}, form two simplices, with \begin{math} S4 \end{math} being a \emph{0}-simplex. The sentences are denoted by colored ellipses, while the high-dimensional embedding of each word in a sentence is depicted by a point having the same color as the parent sentence ellipse. The UMAP algorithm is then employed to find a similarity-preserving Euclidean embedding-space, shown on the right, by minimizing the cross-entropy between the two representations.}
    \label{figdistemb}
\end{figure}

A key difference, in this work, to the original formulation is that an individual data sample (i.e., the vertex of a simplex) is not a \begin{math} d \end{math}-dimensional point but a set of \begin{math} d \end{math}-dimensional words that make up a sentence. By using any of the distance metrics defined in Section \ref{distance}, it is possible to construct the simplicial complex that UMAP needs in order to build the topological representation of the original sentence space. An illustration can be found in Figure \ref{figdistemb}.

As per the formulation laid out for UMAP, the similarity between sentences \begin{math} s' \end{math} and \begin{math} s \end{math} is defined as:
\begin{equation}
    v_{s'|s} = \exp{\frac{-(D(s,s') - \rho_{s})}{\sigma_{s}}}
    \label{orig}
\end{equation}
where \begin{math} \sigma_s \end{math} is a normalisation factor selected based on an empirical heuristic (See Algorithm 3 in the work of \citealt{2018arXivUMAP}), \begin{math} D(s, s') \end{math} is the distance between two sentences as outlined by Equation \ref{ed}, \ref{hdist} or \ref{wmdeq}, and \begin{math} \rho_{s} \end{math} is the distance of \begin{math} s \end{math} from its nearest neighbour. It is worth mentioning that for scalability, \begin{math} v_{s'|s} \end{math} is calculated only for predefined set of approximate nearest neighbours, which is a user-defined input parameter to the UMAP algorithm, using the efficient \emph{nearest-neighbour descent} algorithm \cite{dongetal}.

The similarity depicted in Equation \ref{orig} is asymmetric, and symmetrization is carried out by a fuzzy set union using the probabilistic t-conorm:
\begin{equation}
    v_{ss'} = (v_{s'|s} + v_{s|s'}) - v_{s'|s}v_{s|s'}
    \label{symmetrized}
\end{equation}

As UMAP builds a Vietoris-Rips complex governed by Equation \ref{orig}, it can take advantage of the nerve theorem \cite{nerve}, which makes this construction a homotope of the original topological space. In our case, this implies that we can build a simple nearest neighbours graph from a given corpus of sentences, which has certain guarantees of approximating the original topological space, as defined by the aforementioned distance metrics.

\begin{algorithm}[!th]
    \KwData{A pre-trained word-embeddings matrix, \begin{math} \mathscr{W} \end{math}; a set of sentences, \begin{math} \mathscr{S} \end{math}; desired dimension of the generated sentence-embeddings, \begin{math} d_E \end{math}}
    \KwResult{A set of sentence-embeddings, \begin{math} \{s_E\} \in \mathscr{S}_E \end{math}}

    Calculate the distance matrix for the entire set of sentences, such that the distance between any two sentences is given by Equation \ref{ed}, \ref{hdist} or \ref{wmdeq};
    
    Using this distance matrix, calculate the nearest neighbour graph between all input sentences, given by Equations \ref{orig} and \ref{symmetrized};
    
    Calculate the initial guess for the low dimensional embeddings, \begin{math} \mathscr{S}_E \in \mathbb{R}^{|\mathscr{S}| \times D_E} \end{math}, using the graph laplacian of the original nearest neighbour graph;
    
    Until convergence, minimize the cross-entropy between the two representations (Equation \ref{kldiv}) using stochastic gradient descent;
    
    Return the set of \begin{math} d_E \end{math}-dimensional sentence-embeddings, \begin{math} \mathscr{S}_E \end{math};
    
    \caption{Constructing sentence-\emph{E}mbeddings by \emph{M}anifold \emph{A}pproximation and \emph{P}rojection: \emph{EMAP}}
    \label{distemb1}
\end{algorithm}

The next step is to define a similar nearest neighbours graph in a fixed low-dimensional Euclidean space. Let \begin{math} s_E, s'_E \in \mathbb{R}^{d_E} \end{math} be the corresponding \begin{math} d_E \end{math}-dimensional sentence-embeddings. Then the low dimensional similarities are given by:
\begin{equation}
    w_{ss'} = {(1 + a{{||s_E - s'_E||}_2}^{b}))}^{-1}
    \label{trans}
\end{equation}
where, \begin{math} ||s_E - s'_E|| \end{math} is the Euclidean distance between the \begin{math} d_E \end{math}-dimensional embeddings, and setting \begin{math} a, b \end{math} are input-parameters, set to $1.929$ and $0.791$, respectively, as per the original implementation.

The final step of the process is to optimize the low dimensional representation to have as close a fuzzy topological representation as possible to the original space. UMAP proceeds to do so by minimizing the cross-entropy between the two representations:
\begin{equation}
    C = \sum_{s \neq s'} v_{ss'} \log \frac{v_{ss'}}{w_{ss'}} + (1 - v_{ss'})\log \frac{1 - v_{ss'}}{1 - w_{ss'}}
    \label{kldiv}
\end{equation}
usually done via stochastic gradient descent.

A summary of the proposed process used to produce sentence-embeddings is provided in Algorithm \ref{distemb1}, and pictorially presented in Figure \ref{figdistemb}.

\section{Datasets and resources}
\label{sec:data}

\subsection{Datasets}
\label{sec:dataset}

\begin{table*}[!hbtp]
\small
\centering
\begin{tabular}{|c|c|c|c|c|c|}
\hline
\textbf{Dataset} & \textbf{\#classes} & \textbf{\#train docs} & \textbf{\#test docs} & \textbf{\#avg tokens} & \textbf{Data-details} \\ \hline
\textbf{amazon} & 4 & 5600 & 2400 & 70 & Reviews labeled by product \\ \hline
\textbf{bbcsport} & 5 & 517 & 220 & 192 & Articles labeld by sport \\ \hline
\textbf{classic} & 4 & 4965 & 2128 & 62 & Manuscripts labeled by publisher \\ \hline
\textbf{ohsumed} & 10 & 3999 & 5153 & 104 & Medical abstracts categorized by subject headings \\ \hline
\textbf{reuters8} & 8 & 5485 & 2189 & 69 & News article categorization \\ \hline
\textbf{twitter} & 3 & 2176 & 932 & 8 & Tweet sentiment analysis \\ \hline
\end{tabular}
\caption{\textbf{Dataset information}: Metadata describing the datasets used in our experiments.}
\label{datasets}
\end{table*}

Six public datasets\footnote{\url{https://drive.google.com/open?id=1sGgAo2SBoYKhQQK_kilUp8KSToCI55jl}} have been used to empirically validate the method proposed in this paper. These datasets are of varying sizes, tasks and complexities, and have been used widely in existing literature, thereby making comparisons and reporting possible. Information about the datasets can be found in Table \ref{datasets}.

\subsection{Resources}
\label{resources}

\textbf{Pre-trained word-embedding corpus}: We use the pre-trained set of word-embeddings provided by Mikolov et al \shortcite{w2v}\footnote{\url{https://drive.google.com/file/d/0B7XkCwpI5KDYNlNUTTlSS21pQmM/edit}}.

\noindent\textbf{Software implementations}: We use a variety of software packages and custom-written programs perform our experiments, the starting point being the calculation of sentence-wise distances. We calculate the Hausdorff distance using a directed implementation provided in the \emph{Scipy} python library\footnote{\url{https://docs.scipy.org/doc/scipy/reference/generated/scipy.spatial.distance.directed\_hausdorff.html}}, whereas the energy distance is calculated using \emph{dcor}\footnote{\url{https://dcor.readthedocs.io/en/latest/functions/dcor.energy\_distance.html\#dcor.energy\_distance}}. Lastly, the word mover's distance is calculated using implementation provided by Kusner et al. \shortcite{WMDist}\footnote{\url{https://github.com/mkusner/wmd}}. In order to produce the symmetric distance matrix for a dataset, we employ custom parallel implementation which distributes the calculations over all available logical cores in a machine.

To calculate the sentence-embeddings, the implementation of UMAP provided by McInnes et al \shortcite{mcinnes2018umap-software} is used\footnote{\url{https://umap-learn.readthedocs.io/en/latest/api.html}}. Finally, the classification is done via linear kernel support vector machines from the \emph{scikit-learn} library \cite{Pedregosa:2011:SML:1953048.2078195}\footnote{\url{https://scikit-learn.org/stable/modules/generated/sklearn.svm.SVC.html}}.

All of the code and datasets have been packaged and released\footnote{\url{https://github.com/DeepK/distance-embed}} to rerun all of the experiments.

\noindent\textbf{Compute infrastructure}: All experiments were run on a \emph{m4.2xlarge} machine on AWS-EC2\footnote{\url{https://aws.amazon.com/ec2/}}, which has 8 virtual CPUs and 32GB of RAM.

\section{Experiments}
\label{sec:expresults}

\subsection{Competing methods}
\label{subsec:competing}

In order to check the usefulness of our proposed approach, we benchmark its performance in two different ways. The first, and most obvious, approach is to consider the performance of the \begin{math} k \end{math}-NN classifier as a baseline. This is motivated by the state-of-the-art \begin{math} k \end{math}-NN based classification accuracy reported by \citeauthor{WMDist} for the word mover's distance. Thus, our embeddings need to match or surpass the performance of a \begin{math} k \end{math}-NN based approach, in order to be considered for practical use.

The second approach is to compare the classification accuracies of several state-of-the-art embedding-generation algorithms on our chosen datasets. These are:

\noindent\textbf{dct} \cite{almarwani-etal-2019-efficient}: embeddings are generated by employing discrete cosine transform on a set of word vectors.

\noindent\textbf{eigensent} \cite{kayal-tsatsaronis-2019-eigensent}: sentence representations produced via higher-order dynamic mode decomposition \cite{doi:10.1137/15M1054924} on a sequence of word vectors.

\noindent\textbf{wmovers} \cite{wmemb}: a competing method which can learn sentence representations from the word mover's distance based on kernel learning, termed in the original work as \emph{word mover's embeddings}.

\noindent\textbf{p-means} \cite{pmean}: produces sentence-embeddings by concatenating several power-means of word-embeddings corresponding to a sentence.

\noindent\textbf{doc2vec} \cite{doc2vec}: embeddings produced by jointly learning the representations of sentences, together with words, as a part of the word2vec procedure.

\noindent\textbf{s-bert} \cite{reimers-gurevych-2019-sentence}: embeddings produced by fine-tuning a pre-trained BERT model using a Siamese architecture to classify two sentences as being similar or different.

Note that the results for \emph{wmovers} and \emph{doc2vec} are taken from Table 3 of Wu et al.'s work \shortcite{wmemb}, while all the other algorithms are explicitly tested.

\subsection{Setup}
Extensive experiments are performed to provide a holistic overview of our neighbourhood-preserving embedding algorithm, for various sets of input parameters. The steps involved are as follows:

\noindent\textbf{Choose a dataset} (one of the six mentioned in Section \ref{sec:dataset}). For every word in every sentence in the train and test splits of the dataset, retrieve the corresponding word-embedding from the pre-trained embedding corpus (as stated in Section \ref{resources}).

\noindent\textbf{Calculate symmetric distance matrices} corresponding to each of the chosen distance metrics, for all of the sets of word-embeddings from the train and test splits.

\noindent\textbf{Apply the UMAP algorithm} on the distance matrices to generate embeddings for all sentences in the train and the test splits.

\noindent\textbf{Calculate embeddings for competing methods} for the methods outlined in Section \ref{subsec:competing}.

Embeddings are generated for various hyperparameter combinations for \emph{EMAP} as well as all the compared approaches, as listed in Table \ref{hyperparams}.

\noindent \textbf{Train a classifier on the produced embeddings} to perform the dataset-specific task. In this work, we train a simple linear-kernel support vector machine \cite{Cortes:1995:SN:218919.218929} for every competing method and every dataset tested. The classifier is trained on the train-split of a dataset and evaluated on the test-split. The only parameter tuned for the SVM is the L2 regularization strength, varied between 0.001 and  100. The overall test \emph{accuracy} has been been reported as a measure of performance.

\begin{table*}[!htbp]
\small
\centering
\begin{tabular}{|c|c|c|}
\hline
\textbf{Method}                     & \textbf{Parameter}      & \textbf{Value(s) Tested}                       \\ \hline
\multirow{6}{*}{\textbf{EMAP}}      & \textit{n\_neighbors}   & 40                                             \\ \cline{2-3} 
                                    & \textit{embedding\_dim} & 50, 100, 300, 1000                             \\ \cline{2-3} 
                                    & \textit{min\_dist}      & 1.0, 1.5, 2.0                                  \\ \cline{2-3} 
                                    & \textit{spread}         & 1.0, 2.5                                       \\ \cline{2-3} 
                                    & \textit{n\_iters}       & 1000                                           \\ \cline{2-3} 
                                    & \textit{distance}       & wmd, hausdorff, energy                         \\ \hline
\multirow{2}{*}{\textbf{kNN}}       & \textit{k}              & 1                                              \\ \cline{2-3} 
                                    & \textit{distance}       & wmd, hausdorff, energy                         \\ \hline
\textbf{dct}                        & \textit{components}     & 1 through 6                                    \\ \hline
\multirow{2}{*}{\textbf{eigensent}} & \textit{components}     & 1 through 3                                    \\ \cline{2-3} 
                                    & \textit{time\_lag}      & 1, 2, 3, {[}1,2{]}, {[}1,2,3{]}, {[}1,2,3,4{]} \\ \hline
\textbf{pmeans}                     & \textit{powers}         & 1, {[}1,2{]}, {[}1,2,3{]}, {[}1,2,3,4,5,6{]}   \\ \hline
\textbf{s-bert}                     & \textit{model}          & bert-base-nli-mean-tokens                      \\ \hline
\end{tabular}
\caption{\textbf{Hyperparameter values tested}. For \emph{EMAP}, \emph{n\_neighbours} refers to the size of local neighborhood used for manifold approximation, \emph{embedding\_dim} is the fixed dimensionality of the generated sentence-embeddings, \emph{min\_dist} is the minimum distance apart that points are allowed to be in the low dimensional representation, \emph{spread} determines the scale at which embedded points will be spread out, \emph{n\_iters} is the number of iterations that the UMAP algorithm is allowed to run, and finally, \emph{distance} is one of the metrics proposed in Section \ref{distance}. For the spectral decomposition based algorithms, \emph{dct} and \emph{eigensent}, \emph{components} represents the number of components to keep in the resulting decomposition, while \emph{time\_lag} corresponds to the window-length in the dynamic mode decomposition process. For \emph{pmeans}, \emph{powers} represents the different powers which are used to generate the concatenated embeddings.}
\label{hyperparams}
\end{table*}

\section{Results and Discussion}
\label{results}

The results of all our experiments are in compiled in Tables \ref{versusknn} and \ref{versussota}. All statistical tests reported are z-tests, where we compute the right-tailed p-value and call a result significantly different if $p<0.1$.

\begin{table*}[!t]
\small
\centering
\begin{tabular}{|c|c|c|c|c|c|c|}
\hline
\textbf{Distance} & \multicolumn{2}{c|}{\textbf{energydist}} & \multicolumn{2}{c|}{\textbf{hausdorffdist}} & \multicolumn{2}{c|}{\textbf{wmddist}} \\ \hline
\textbf{Method}   & \textbf{knn}      & \textbf{EMAP}    & \textbf{knn}        & \textbf{EMAP}     & \textbf{knn}    & \textbf{EMAP}   \\ \hline
\textbf{amazon}   & \textbf{0.923}*    & 0.909                & 0.781               & \textbf{0.844}*        & 0.918           & \textbf{0.929}*      \\ \hline
\textbf{bbcsport} & 0.941             & \textbf{0.942}       & 0.925               & \textbf{0.941}        & 0.972           & \textbf{0.987}      \\ \hline
\textbf{classic}  & 0.912             & \textbf{0.921}       & 0.943               & \textbf{0.953}*        & 0.961           & \textbf{0.978}*      \\ \hline
\textbf{ohsumed}  & 0.456             & \textbf{0.505}*       & 0.491               & \textbf{0.603}*        & 0.551           & \textbf{0.630}*      \\ \hline
\textbf{r8}       & 0.942             & \textbf{0.962}*       & \textbf{0.863}*      & 0.837                 & 0.951           & \textbf{0.973}*      \\ \hline
\textbf{twitter}  & 0.731             & \textbf{0.749}       & 0.736               & \textbf{0.741}        & 0.712           & \textbf{0.722}      \\ \hline
\end{tabular}
\caption{\textbf{Comparison versus kNN}. Results shown here compare the classification accuracies of \begin{math}k\end{math}-nearest neighbour to our proposed approach for various distance metrics. For every distance, \textbf{bold} indicates better accuracy, while $*$ indicates that the winning accuracy was statistically significant with respect to the compared value (,i.e., EMAP vs kNN for a given distance metric). It can be observed that our method almost always outperforms \begin{math}k\end{math}-nearest neighbour-based classification.}
\label{versusknn}
\end{table*}

\begin{table*}[!t]
\small
\centering
\begin{tabular}{|c|c|c|c|c|c|c|c|}
\hline
\textbf{Method}   & \textbf{wmd-EMAP} & \textbf{dct} & \textbf{eigensent} & \textbf{wmovers} & \textbf{pmeans} & \textbf{doc2vec} & \textbf{s-bert} \\ \hline
\textbf{amazon}   & 0.929                & 0.932        & 0.902$\vee$              & \textbf{0.943}$\wedge$   & \textit{0.938}      & 0.912$\vee$      & 0.923            \\ \hline
\textbf{bbcsport} & \textbf{0.986}       & 0.972        & 0.968              & \textit{0.982}   & 0.981               & 0.979      & \textbf{0.986}            \\ \hline
\textbf{classic}  & \textbf{0.978}       & 0.964        & 0.947$\vee$              & \textit{0.971}   & 0.960               & 0.965      & 0.966            \\ \hline
\textbf{ohsumed}  & \textit{0.630}       & 0.594$\vee$        & 0.574$\vee$              & \textbf{0.645}$\wedge$   & 0.614$\vee$               & 0.598$\vee$      & 0.556$\vee$            \\ \hline
\textbf{r8}       & \textbf{0.973}       & 0.967        & 0.958$\vee$              & \textit{0.972}   & 0.969               & 0.949$\vee$      & 0.954$\vee$            \\ \hline
\textbf{twitter}  & \textit{0.722}       & 0.644$\vee$        & 0.669$\vee$              & \textbf{0.745}   & 0.636$\vee$               & 0.673$\vee$      & 0.673$\vee$            \\ \hline
\end{tabular}
\caption{\textbf{Comparison versus competing methods}. We compare \emph{EMAP} based on word mover's distance to various state-of-the-art approaches. The best and second-best classification accuracies are highlighted in \textbf{bold} and \textit{italics}. We perform statistical significance tests of our method (\emph{wmd-EMAP}) against all other methods, for a given dataset, and denote the outcomes by $\vee$ when the compared method is worse and $\wedge$ when our method is worse, while the absence of a symbol indicates insignificant differences. In terms of absolute accuracy, we observe that our method achieves state-of-the-art results in 2 out of 6 datasets.}
\label{versussota}
\end{table*}

\begin{table*}[!t]
\small
\centering
\begin{tabular}{|l|l|c|}
\hline
\multicolumn{1}{|c|}{\textbf{Query Sentence}}                                                                                                                                                                                                                                                                                                                                                                                                                                                                                                                                                                                                                                                                                                                                                                                                                                                                                                                                                                                                                                                                                                                                                                                                                                                                                                              & \multicolumn{1}{c|}{\textbf{Best Match Sentence}}                                                                                                                                                                                                                                                                                                                                                                                                                                                                                                                                                                                                 & \textbf{Cosine Sim} \\ \hline
\begin{tabular}[c]{@{}l@{}}I have spent thousands of dollar’s On Meyers\\ cookware everthing from KitchenAid Anolon\\ Prestige Faberware \& Circulan just to name a few\\ Though Meyers does manufacture very high quality\\ pots \& pans and I would recommend them to anyone\\ it’s just sad that if you have any problem with them\\ under warranty you have to go throught the chain\\ of command that never gets you anywhere even if\\ you want to speak with upper management about\\ the rudeness of the customer service department\\ Their customer service department employees are\\ always very rude and snotty and they act like they\\ are doing you a favor to even talk to you about their\\ products\end{tabular}                                                                                                                                                                                                                                                                                                                                                                                                                                                                                                                                                                                                                        & \begin{tabular}[c]{@{}l@{}}When I opened the box I noticed corrosion\\ on the lid When I contacted Rival customer\\ service via email they told me I had to purchase\\ a new lid I called and spoke with a customer\\ service representative and they told me that a\\ lid was not covered under warranty When I\\ explained that I just opened it and it was\\ defective they told me to just return the\\ product that there was nothing that they were\\ going to do After being treated this way I will\\ NOT be purchasing any more Rival products\\ if they don’t stand behind their product VERY\\ VERY poor customer service\end{tabular} & 0.997               \\ \hline
\begin{tabular}[c]{@{}l@{}}This movie will bring up your racial prejudices in\\ ways that most movies just elude to It demonstrates\\ how connected we all are as people and how seperated\\ we are by only one thing our viewpoints The acting\\ is superb and you get one cameo appearance after\\ another which is a treat Of course the soundtrack is\\ terrific The ending is intense to witness one situation\\ after another coming to an unfortunate finish\end{tabular}                                                                                                                                                                                                                                                                                                                                                                                                                                                                                                                                                                                                                                                                                                                                                                                                                                                                           & \begin{tabular}[c]{@{}l@{}}I waited years for this movie to be released in the\\ United States As far as I was concerned it wasn’t\\ about the acting as much as it was about the\\ feeling the actors wanted to portray in which\\ they profoundly accomplished I would recommend\\ this movie to anyone who can reach that one step\\ deeper into the minds of creativity and passion\\ and appreciate the struggles of rising above and\\ beyond the pain of broken dreams\end{tabular}                                                                                                                                                        & 0.998               \\ \hline
\begin{tabular}[c]{@{}l@{}}We see a phrase a lot when we visit how to sites for \\ writers World building By this we mean the setting\\  the characters and everything else where our story\\  will occur For me this often means maps memories\\  and visits since I write about where I live But if\\  you'd like to see exactly what world building means\\  head down to your local library and grab SALEM'S\\  LOT by Stephen King When Stephen \\ King mania first gripped the English speaking world\\  I missed it I saw the film of CARRIE and hated it \\ Years later at a guard desk on a long shift scheduled\\  so suddenly that I hadn't had a chance to visit the \\ library I read what was in the desk instead THINNER\\ If I were Stephen King I'd have put a pen name on\\ that crap as well One of King's fans brought me \\ around She recommended THE SHINING Of course\\ I thought of that Kubrick/Nicholson travesty No no \\ she said read the book It's much different Yes it is \\ It's fantastic for its perceptiveness Next up PET \\ SEMATARY which scared the crap out of me \\ And that my friends is not easy ON WRITING I've\\ gushed about that enough times The films STAND \\ BY ME and THE APT PUPIL So in the end I \\ appreciate King and forgive him for CARRIE \\ and I think he's forgiven himself\end{tabular} & \begin{tabular}[c]{@{}l@{}}in the possibility that Steve Berry could ever \\ transcend his not so great debut The Amber Room\\ Romanov Prophecy started in the right direction\\ Third Secret was OK but I think he hit his *peak*\\ right there\end{tabular}                                                                                                                                                                                                                                                                                                                                                                                     & 0.955               \\ \hline
\end{tabular}
\caption{\textbf{Examples of best-matching sentences}. From the \emph{amazon} reviews dataset using \emph{wmd-EMAP}.}
\label{sentcomparison}
\end{table*}

\noindent\textbf{Performance of the distance metrics}: From Table \ref{versusknn} it can be observed that the word mover's distance consistently performs better than the others experimented with in this paper. WMD calculates the total effort of aligning two sentences, which seems to capture more useful information compared to the hausdorff metric's worst-case effort of alignment. As for the energy distance, it calculates pairwise potentials amongst words within and between sentences, and may suffer if there are shared commonly-occurring words in both the sentences. However, given that energy and hausdorff distances are reasonably fast to calculate and perform respectably well, they might be worth using in applications with a large number of long sentences.

\noindent\textbf{Comparison versus kNN}: \emph{EMAP} almost always outperforms \begin{math}k\end{math}-nearest neighbours based classification, for all the tested distance metrics. The performance boost for WMD is between a relative percentage accuracy of 0.5\% to 14\%. This illustrates the efficiency of the proposed manifold-learning method.

\noindent\textbf{Comparison versus state-of-the-art methods}: Consulting Table \ref{versussota}, it seems that \emph{wmovers}, \emph{pmeans} and \emph{s-bert} form the strongest baselines as compared to our method, \emph{wmd-EMAP} (\emph{EMAP} with word mover's distance). Considering the statistical significance of the differences in performance between \emph{wmd-EMAP} and the others, it can be seen that it is almost always equivalent to or better than the other state-of-the-art approaches. In terms of absolute accuracy, it \emph{wins} in 3 out of 6 evaluations, where it has the highest classification accuracy, and comes out second-best for the others. Compared to it's closest competitor, the \emph{word mover's embedding} algorithm, the performance of \emph{wmd-EMAP} is found to be on-par (or slightly better, by 0.8\% in the case of the \emph{classic} dataset) to slightly worse (3\% relative p.p., in case of the \emph{twitter} dataset). Interestingly, both of the distance-based embedding approaches, \emph{wmd-EMAP} and \emph{wmovers}, are found to perform better than the siamese-BERT based approach, \emph{s-bert}.

Thus, the overall conclusion from our empirical studies is that \emph{EMAP} performs favourably as compared to various state-of-the-art approaches.

\noindent\textbf{Examples of similar sentences with EMAP}: We provide motivating examples of similar sentences from the \emph{amazon} dataset, as deemed by our approach, in Table \ref{sentcomparison}. As can be seen, our method performs quite well in matching complex sentences with varying topics and sentiments to their closest pairs. The first example pair has the theme of a customer who is unhappy about poor customer service in the context of cookware warranty, while the second one is about positive reviews of deeply-moving movies. The third example, about book reviews, is particularly interesting: in the first example, a reviewer is talking about how she disliked the first Stephen King work which she was exposed to, but subsequently liked all the next ones, while in the matched sentence the reviewer talks about a similar sentiment change towards the works of another author, Steve Berry. Thus in the last example, the similarity between sentences is the change of sentiment, from negative to positive, towards the works of books of particular authors.

\section{Conclusions}
\label{sec:conclusions}

In this work, we propose a novel mechanism to construct unsupervised sentence-embeddings by preserving properties of local neighbourhoods in the original space, as delineated by set-distance metrics. This method, which we term, \emph{EMAP} or \emph{Embeddings by Manifold Approximation and Projection} leverages a method from topological data analysis can be used as a framework with any distance metric that can discriminate between sets, three of which we test in this paper. Using both quantitative empirical studies, where we compare with state-of-the-art approaches, and qualitative probing, where we retrieve similar sentences based on our generated embeddings, we illustrate the efficiency of our proposed approach to be on-par or exceeding in-use methods. This work demonstrates the successful application of topological data analysis in sentence embedding creation, and we leave the design of better distance metrics and manifold approximation algorithms, particularly targeted towards NLP, for future research.

\bibliography{eacl2021}
\bibliographystyle{acl_natbib}

\end{document}